\documentclass[]{article}

\usepackage[left=2cm, right=2cm, top=2cm, bottom=2cm]{geometry}
\usepackage{graphicx}
\usepackage{subcaption}
\usepackage{hyperref}

\usepackage{amssymb}
\usepackage{latexsym}

\usepackage{framed,multirow}

\usepackage{url}
\usepackage{xcolor}
\definecolor{newcolor}{rgb}{.8,.349,.1}

\title{A Comparative Study of Face Detection Algorithms for Masked Face Detection}

\author{
  Iqbal, Sahel Mohammad\\
  \texttt{sahelm.iqbal@niser.ac.in}
  \and
  Shekar, Danush\\
  \texttt{s.danush@yahoo.com}
  \and
  Mishra, Subhankar\\
  \texttt{smishra@niser.ac.in}
}

\begin{document}

\maketitle

\begin{abstract}
Contemporary face detection algorithms have to deal with many challenges such as variations in pose, illumination, and scale. A subclass of the face detection problem that has recently gained increasing attention is occluded face detection, or more specifically, the detection of masked faces. Three years on since the advent of the COVID-19 pandemic, there is still a complete lack of evidence regarding how well existing face detection algorithms perform on masked faces. This article first offers a brief review of state-of-the-art face detectors and detectors made for the masked face problem, along with a review of the existing masked face datasets. We evaluate and compare the performances of a well-representative set of face detectors at masked face detection and conclude with a discussion on the possible contributing factors to their performance.
\end{abstract}


\section{Introduction}

The development of machine learning algorithms has directed a line of research towards object-detection algorithms, of which face detection algorithms form a subset. Recent years have witnessed a remarkable increase in the performance of these algorithms in terms of precision and speed. There has been significant interest in the field because face detection algorithms have broad applications like biometric security, surveillance, and video indexing. Faces vary in age, gender, size, pose, illumination, and occlusion, and a good face detector must be able to deal with all of these variations. Although there has been considerable progress in tackling such issues (Ref. \cite{pose} for pose estimation, Single-Shot Scale Invariant Face Detector (S3FD) \cite{s3fd} to have scale-invariant performance, and FaceBoxes \cite{faceboxes} to perform at faster speeds), there are yet not many models that tackle the occlusion problem specifically.

During the COVID-19 pandemic, the majority of world nations rolled out mask mandates. Three years on from the start of the pandemic, wearing a mask in public places is still a requirement for citizens in some countries. This has forced many industries to modify existing technology, of which a relatable example is the mobile phone industry slowly moving to make their face recognition-based screen locks more adapted to faces with masks. However, in most areas of application/industries, people are still employing the same face detection and recognition algorithms as they were pre-pandemic. There is currently a complete absence of objective evidence on how well current systems perform on masked faces, which may be problematic, particularly in areas where detection systems need to be highly robust. Thus, to quantify the reliability and accuracy of current systems under this novel scenario, it is crucial to understand and study how well these standard face detectors can perform when dealing with masked faces.

\begin{figure}
    \centering
    \includegraphics[width=3in]{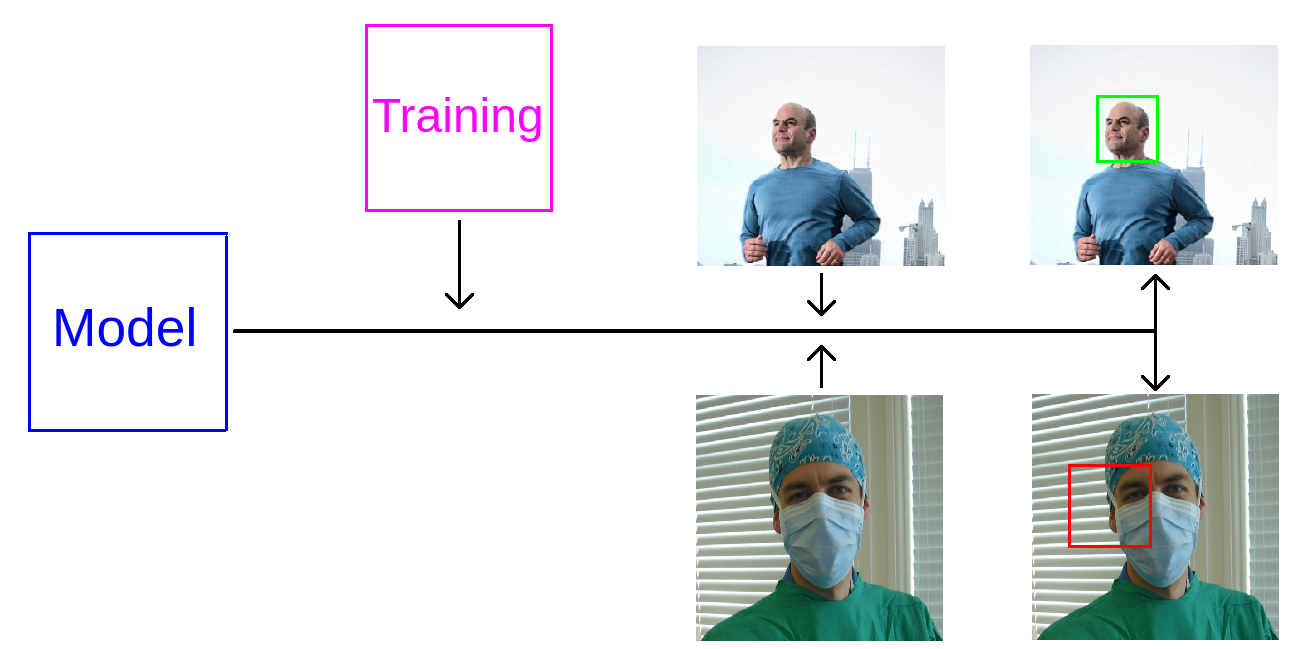}
    \caption{A schematic image depicting how current models that have undergone training with current datasets, might not perform well on images with masked faces.}
    \label{fig:status}
\end{figure}

Numerous comprehensive surveys like Ref. \cite{occ-survey-1}, \cite{occ-survey-2} and \cite{occ-survey-3} have been written to tackle the occluded face detection and recognition problem. Compared to face recognition, there are fewer surveys for face detection, and to the extent of our knowledge, there are no surveys for masked face detection. Ref. \cite{occ-survey-1} in particular, only touches upon the task of masked face detection while introducing the MAsked FAces (MAFA) dataset \cite{mafa}, and is more aligned to dealing with occlusion in general. This might be because there are very few algorithms made specifically for masked face detection, and thus makes it too early to survey face detectors that tackle the masked face problem. To the extent of our knowledge, there are also no prior surveys that address masked face detection accuracy by standard face detection models. 


In this work, we seek to fill this void by evaluating the performance of multiple state-of-the-art face detectors at the task of masked face detection. The objectives of this work are:
\begin{itemize}
    \item Review state-of-the-art face detectors, including those that are made to detect occluded and masked faces.
    \item Review existing masked face datasets and what they lack, and in our case, simulate a new masked face dataset for reasons that will be explained later.
    \item Evaluate the performance of a representative set of face detectors at masked face detection. This study would help research shine light on directions that show promising performance in the task of masked face detection.
    \item Limitations and strengths of the face detectors that we tested.
\end{itemize}

The remainder of the article is organized as follows: Section \ref{sec:survey} presents the reader with a brief survey of face detection models, datasets for occluded and masked faces, and models designed specifically for detecting occluded faces. Our research methodology is presented in Section \ref{sec:methodology}, which details the criteria for selecting models for our study and how they were trained and tested. In Section \ref{sec:implementationdetails} we elaborate on the implementation details of the various models and what makes them different from one another. In section \ref{sec:results} we present the results of testing our models on a custom masked dataset we created using WIDER FACE \cite{widerface} images. We also present discussions on why some models perform better than others. Finally, Section \ref{sec:conclusions} concludes this article with concluding remarks, the limitations of the models we tested and that of our testing methodology.


\section{Masked Face Detection}

Masked face detection forms a sub-class of the face detection problem, wherein the latter is used for the task of detecting faces (and distinguishing faces amongst other objects) in an image or video. Sub-tasks of face detection include problems like pose estimation, occluded face detection, etc. Masked face detection also comes under the task of face detection, and refers to the detection of masked faces. As explained earlier, this problem has become more relevant in recent years owing to the COVID-19 pandemic that has shaped the development of technologies that were dependent on face detection methods to also improve masked face detection.

\begin{figure}[h]
    \centering
    \includegraphics[width =3in]{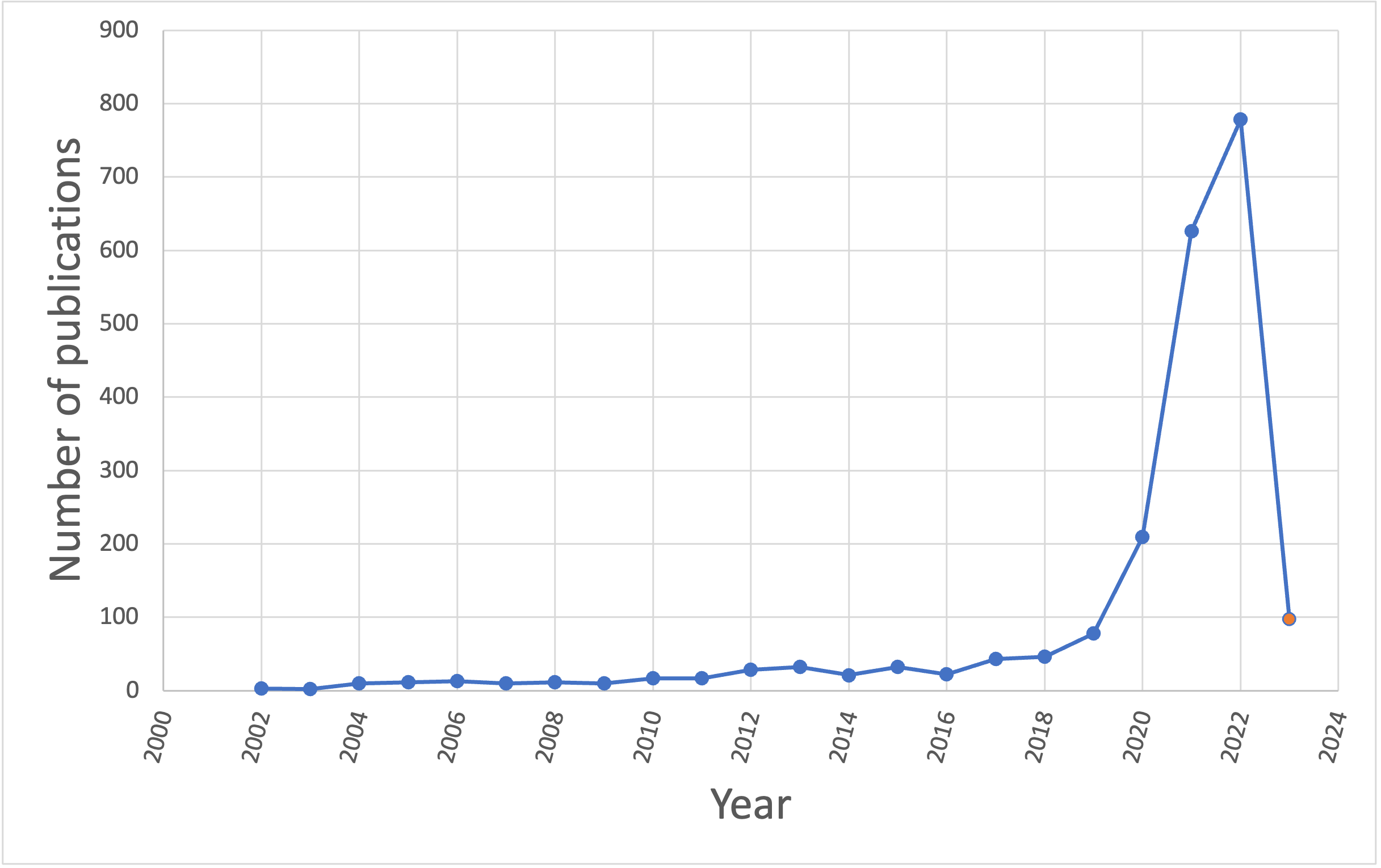}
    \caption{Number of papers published in the last 2 decades related to masked face detection. Image obtained from \href{https://app.dimensions.ai}{https://app.dimensions.ai} on March 2023 with the search keyword ‘‘mask’’, ‘‘face’’, and ‘‘detection’’ applied to the title and abstract of scholarly papers. The current year's data point is in orange to signify that the point can vary as more articles are published this year.}
    \label{fig:biblio}
\end{figure}

Compared to the masked face detection, the occluded face detection problem is relatively older, as occlusion form a superset of masked faces, and images of occluded faces have had more occurrences in images than masked faces (which has up-risen only in the recent times) since the advent of the face detection problem. Additionally, with data obtained from \href{https://app.dimensions.ai}{https://app.dimensions.ai} (See Fig. \ref{fig:biblio}), we can infer that the research done in this sub-topic is in its early stages and that the number of papers published every year is rising significantly.

\section{Literature Survey}\label{sec:survey}

One of the earliest face detection algorithms was Viola-Jones \cite{violajones}, which used Haar-like features and the AdaBoost algorithm to train a cascade of classifiers. The Deformable Part Model (DPM) \cite{dpm} then rose as a popular object detection method with an extension of its use to face detection \cite{dpm-with-face-detection1, pose, dpm-with-face-detection3, dpm-with-face-detection4}. Algorithms based on DPMs were limited by their speed, that said, 
face detectors have come a long way since then after the development of Convolutional Neural Networks (CNNs). Right now, there are a plethora of face detectors and they are truly multifaceted, due to which there aren't many classification schemes to separate them all into disjoint classes. Similar to \cite{retinaface} and \cite{centerface}, we categorize face detection models into the following classes:

\textbf{Anchor-based vs Anchor-free algorithms}: Most algorithms use anchor boxes of multiple ratios and sizes (although a square anchor is most commonly used) to detect faces and these anchors are usually predefined. Algorithms like PyramidBox \cite{pyramidbox}, TinyFaces \cite{tinyfaces}, RetinaFace \cite{retinaface}, and FaceBoxes \cite{faceboxes} use anchor-based methods to perform the task of face detection. Even though anchor-based methods are robust and achieve speeds that are invariant to the number of objects they are required to detect, they have some issues like the drop in accuracy with tiny faces \cite{s3fd, anchor-issues}. To this effect algorithms like the Light and Fast Face Detector (LFFD) \cite{lffd} and CenterFace \cite{centerface} have proposed anchor-free methods.

\textbf{Single-stage vs Two-stage models}: Single-stage algorithms like PyramidBox \cite{pyramidbox}, S3FD \cite{s3fd} (which is also a scale-invariant algorithm) and RetinaFace \cite{retinaface} densely sample face locations and scales, on feature pyramids. Two-stage designs involve a proposal and a refinement stage, and algorithms like CMS-RCNN \cite{cms-rcnn}, Faster-R-CNN \cite{faster-r-cnn}, and Selective Refinement Network (SRN) \cite{srn} employ a two-stage design. Single-stage designs are faster compared to their two-stage counterparts but may perform worse in terms of accuracy.


\textbf{Context-associated methods}: Faces almost never appear as a sole entity, other regions of the body like the shoulders and the neck are usually associated with each face in an image. Some algorithms exploit this property and utilize contextual information to better locate faces, and some algorithms that employ such methods include PyramidBox \cite{pyramidbox}, CMS-RCNN \cite{cms-rcnn}, FaceBoxes \cite{faceboxes}, SSH\cite{ssh}, and Tiny Faces \cite{tinyfaces}.

\textbf{Multitask learning}: Another category for classifying face detectors is one that use a combination of  two tasks with the assumption/knowledge that there is an inherent correlation between the two tasks. For example, the detection and alignment tasks are combined to give joint face detection and alignment algorithms like the Multitask Cascaded Convolutional Networks (MTCCN) \cite{mtccn} and the Joint Cascade Face Detection and Alignment \cite{joint-cascade-face-detection-and-alignment}. Another example of multitask learning is the Mask R-CNN \cite{mask-r-cnn} algorithm which significantly improves the detection performance by adding a branch for predicting segmentation masks on every region of interest in conjunction with the existing branch for classification and regression.

With respect to occluded face detection, and particularly for masked faces, the number of scientific articles and models is far less compared to the task of face detection. There is still a lot of progress in performance needed in this direction \cite{masked-face-review-deep-learning, NIST-review}. The LLE-CNN is an algorithm to detect masked faces that was introduced along with the same article that introduced the MAFA dataset \cite{mafa}. Face Attention Network (FAN) \cite{fan} uses a feature pyramid network with settings similar to that of RetinaNet \cite{ref26}, made for detecting occluded faces. FAN has been evaluated on the WIDER Face and MAFA datasets showing good accuracies compared to other occluded face detection algorithms like Adversarial Occlusion-aware Face Detection (AOFD) \cite{aofd} and LLE-CNNs. AOFD is another algorithm made to detect occluded faces using adversarial networks, and the model treats occluded regions as an auxiliary feature to detect faces. Ref. \cite{grid-loss} is another occluded face detector that has a grid loss layer embedded into CNNs that minimized the error rates on each sub-block of the feature map independently, but it should be noted that testing on masked datasets was not performed by the authors. Ref. \cite{dpm-occ-model} also includes work done towards synthesizing new images for training their model, because existing datasets primarily contain images with fully visible faces. There are other algorithms such as FaceMaskNet-21 \cite{facemasknet-21}, RetinaFaceMask \cite{retinamask}, and Ref. \cite{masked-face-recognition} that have conducted studies on masked faces. FaceMaskNet-21 is a recent algorithm for detecting masked faces, trained primarily with the LFW \cite{lfw} dataset, and also a dataset of 204 images compiled by the authors themselves. RetinaFaceMask is an algorithm that detects the presence of a face-mask and has been tested on the Face Mask Dataset (\href{https://github.com/AIZOOTech/FaceMaskDetection}{FMD}) \cite{face-mask-detection} which contains only 7971 images. Ref. \cite{masked-face-recognition} conducted studies using the RMFRD and SMFRD datasets. There are older works on occluded faces (and not masked faces specifically) like Ref. \cite{dpm-occ-model} that use a DPM. The authors of this article demonstrate how a deformable template model distorts predictions in the presence of occlusions, and they go on to propose a model that localizes on occluded parts to produce better results. Deep Convolutional Network-based works (like Ref. \cite{deep-learning-occ-model}) for such tasks also exist.

\section{Research Methodology}\label{sec:methodology}
\subsection{Face Detection algorithms}

The decision of which models to choose for the survey was primarily based on two considerations. The first was to encompass as many varieties in architecture and implementation details as possible. Every face detection algorithm has unique features, and our goal was to find out which of those would give an advantage in detecting masked faces. We also wanted to include models that varied in size and complexity, ranging from state-of-the-art models with a large number of parameters to smaller models which could be run on mobile devices and cameras. The second was a practical consideration regarding the models we had access to. We could only attempt to run models whose code has been made public by their authors.

Based on the above two criteria, we chose the following face detection models - MTCCN \cite{mtccn}, TinyFaces \cite{tinyfaces}, FaceBoxes \cite{faceboxes}, RetinaFace \cite{retinaface}, Extremely Tiny Face Detector (EXTD) \cite{extd}, LFFD \cite{lffd}, and CenterFace \cite{centerface}. These models span a large variety of algorithms, with the oldest among the bunch, MTCCN, being introduced in 2016 while the most recent one, CenterFace, was released in 2019. RetinaFace \cite{retinaface} is one of the most accurate face detection models boasting an accuracy of $91.4\%$ on the WIDER Face test (Hard) dataset. CenterFace is also a model that shows impressive accuracy \cite{centerface}. These models were selected because we wanted to see how some of the top-performing face detection models would fare at the task of masked face detection. We chose EXTD and LFFD because they have a minimal number of parameters and have been specifically built to perform inference on devices with limited computational power such as mobile phones, security cameras, etc, and hence are the ones with the greatest potential for widespread adoption.

\begin{table}[htpb]
\centering
\begin{tabular}{|p{2cm}|p{3.25cm}|p{2cm}|}
\hline
Model      & Reported accuracy on WIDER Face test Hard & Year introduced \\ \hline
RetinaFace & 0.914                                                                               & 2019            \\ \hline
CenterFace & 0.873                                                                               & 2019            \\ \hline
EXTD       & 0.850                                                                               & 2019            \\ \hline
TinyFaces  & 0.831                                                                               & 2017            \\ \hline
LFFD       & 0.770                                                                               & 2019            \\ \hline
MTCCN      & 0.607                                                                               & 2016            \\ \hline
FaceBoxes  & 0.396                                                                               & 2017            \\ \hline
\end{tabular}%
\caption{\label{tab:models-tested}Models that were tested as described in Section \ref{sec:methodology}.}
\end{table}

\subsection{Datasets}

\subsubsection{Training}

Since the goal of this work is to study how currently deployed face detection algorithms perform at masked face detection, the models should be trained on standard, un-masked datasets. We used pre-trained weights provided by the models' authors for all the architectures that we tested. We chose to do this instead of training the models on our own for two reasons. Firstly, we wanted to reproduce the exact same inference accuracies reported by the models' authors. Secondly, as our goal was to understand how these models performed without any modification to their training or implementation procedures, training these models from scratch was unnecessary. All the models that we tested were trained either on the WIDER Face or FDDB datasets. WIDER Face is a standard dataset used to train and test face detection models with 32,203 images and 393,703 faces split across training, validation, and testings sets \cite{widerface}, which has a good representation of faces in the wild. The validation and test datasets are further split into three subsets called Easy, Medium, and Hard, with the Hard set having a disproportionately large amount of small and occluded faces.

\subsubsection{Testing}

To test the models, we needed a dataset of masked faces. However, the current selection of masked datasets is limited both in terms of their quality and quantity. The MAFA dataset is currently one of the few datasets that exist for masked faces, along with other datasets such as Masked Face Detection Dataset (MFDD), Real-world Masked Face Recognition Dataset (RMFRD), and Simulated Masked Face Recognition Dataset (SMFRD) \cite{smfrd}. MAFA contains 30,811 images and 34,806 masked faces, but Ref. \cite{retinamask} has described how this dataset contains some faces occluded by hands or other objects rather than physical masks. MFDD, RMFRD, and SMFRD are from the same article \cite{smfrd}, and each of them has certain drawbacks. RMFRD contains 95,000 images, but only 5,000 of these images have masked faces. Among these images, there are only 525 unique subjects, which hint at lesser variation and a disproportion of masked and unmasked faces. SMFRD contains images from face datasets like Webface \cite{webface} and LFW \cite{lfw}, where the authors simulated masked faces by placing masks on these faces. In some ways, this is similar to the process done in this work, except that we have performed this on the WIDER Face dataset, which is widely considered as a benchmark to test on, compared to Webface \cite{webface} and LFW \cite{lfw}. For this reason, we have not used the SMFRD dataset. Although MFDD contains 24,771 masked face images, there have not been many works that have tested this dataset (including Ref. \cite{masked-face-recognition} which tested SMFRD and RMFRD, but not MFDD) and there is simply not enough information on the quality of this dataset. For similar reasons, we dropped the MAFA dataset as well, although future work may be done using this dataset. As introduced earlier, there is FMD, but it contains only 7971 images, and this is not a sufficient dataset size to work with. 

Aside from the existing masked datasets that were mentioned earlier, there has been development in producing good quality synthetic images of masked faces through unique ways like deep learning models. Among them, the popular tools are  MaskedFace-Net \cite{maskedface-net}, MaskTheFace \cite{deep-learn-synthetic-dataset-1}, CelebFaces Attributes (CelebA) dataset \cite{celeba}, Masked Face Segmentation and Recognition (MSFR) dataset \cite{mfsr}, and SMFRD \cite{smfrd}.

The CelebA dataset has more than 200k face images of celebrities with various attributes labeled. The Synthetic CelebA dataset \cite{synthetic-celeba} was built by superimposing masks (randomly chosen from a set of 50 different masks that vary by their color, shape, size, etc) onto the images from CelebA. The MSFR has a combination of images collected from both the internet and the real-world. It has two subsets, the first of which has images of masked faces that have segmentation information labeled, while in the second set, for each unique individual present in the dataset, there is at least one masked and one unmasked image. The subsets have 9742 and 11615 images respectively. Another dataset that can be considered for evaluation is the Masked Faces in Real World for Face Recognition (MFR2) \cite{mfr2}. It has 269 images of 53 individuals which were collected from the internet. There are both masked and unmasked faces of each individual. Reference \cite{mfr2} also introduces LFW-SM (LFW with Simulated Masks), which is a dataset with simulates masks on images generated from the LFW dataset. LFW-SM has 13233 images of 5749 individuals.

The tools discussed above have all made progress towards tackling the masked face detection problem, but each tool has its set of advantages and disadvantages. Since they run a model to detect the faces before editing the images to attach the masks, they are not always accurate. There is a lack of variety in the types of masks, and they are also associated with the problem of masks not matching the tone and color scheme of the image by eclipsing faces, in terms of clarity and brightness. That said, these methods of producing synthetic images are still in their early stages, and it is early to comment and let alone test on such datasets.

\begin{table*}[ht]
\centering
\begin{tabular}{|p{2.5cm}|p{1.5cm}|p{4cm}|p{6cm}|}
\hline
Dataset & Dataset Size & Range & Additional comments  \\ \hline
MAFA  & 30,811  & Masked and occluded faces & Contains some faces occluded by hands or other objects rather than physical masks  \\ \hline
RMFRD & 95,000  & Masked and unmasked faces & Disproportionate amount of unmasked faces compared to masked faces \\ \hline
SMFRD & 500,000 & Simulated masked faces from the Webface and LFW datasets & Simulated dataset, not from an established dataset \\ \hline
MFDD  & 24,771  & Masked faces & Not tested by many models \\ \hline
Maskedface-Net & 137,016 & Masked and unmasked faces & Created with the objective of detecting masks, hence contains a good number of masked and unmasked faces \\ \hline
Face Mask \newline Detection & 7971  & Masked faces & Dataset size is too small \\ \hline
\hline
\end{tabular}
\caption{\label{tab:dataset-summary}A summary of a few datasets with masked faces that were reviewed, and some details and comments on the same.}
\end{table*}

As evidenced in the discussion above, at the time of performing this survey, among other reasons stated earlier, there existed no masked dataset that is genuinely representative of the large variety of masked faces that we see in the wild. As a result, for testing, we decided to simulate a masked dataset by blackening out the lower half of all faces in the WIDER Face validation dataset. This way, the models would have to detect faces based on the information from the upper half of faces only, with the idea that this technique may work regardless of the type of mask used to obscure the rest of the face. Also, since the models are only tested and not trained on this modified dataset, we can rule out the possibility that the models are identifying faces based on rectangular black blobs. Another advantage of pursuing this direction revolves around the certainty that the blacked-out region will be on the lower half of the face, whereas the simulated datasets have chances of adding masks in wrongly identified face regions and also faces with no masks also.

\begin{figure}[!h]
    \centering
    \includegraphics[width=2in]{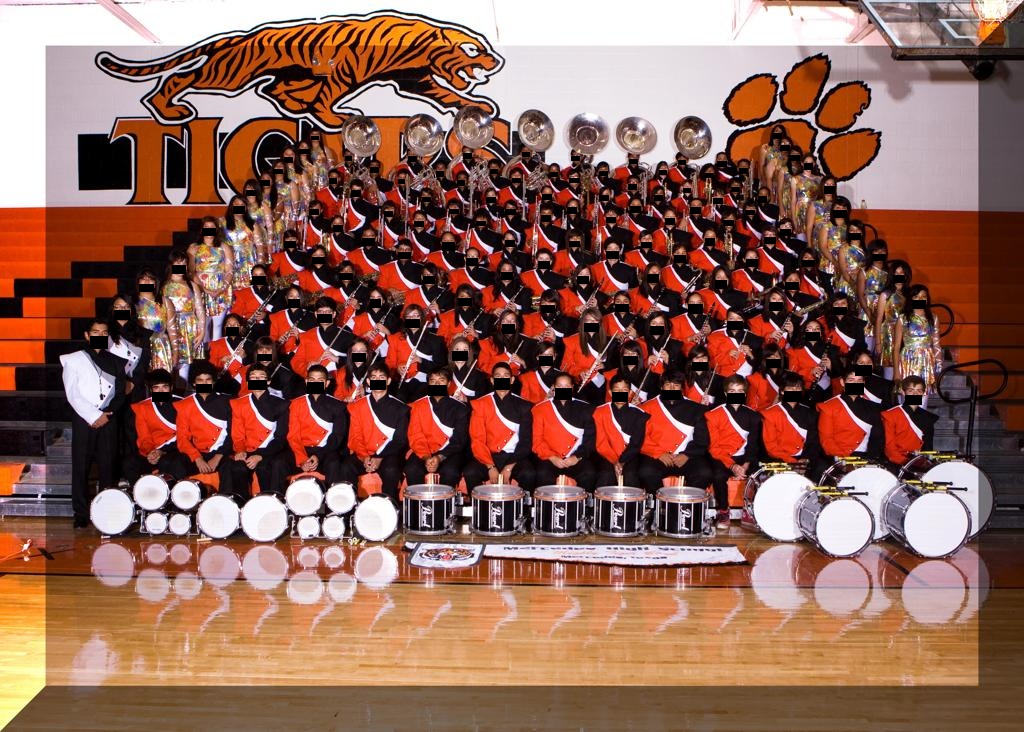}
    \vskip\baselineskip
    \includegraphics[height=2in]{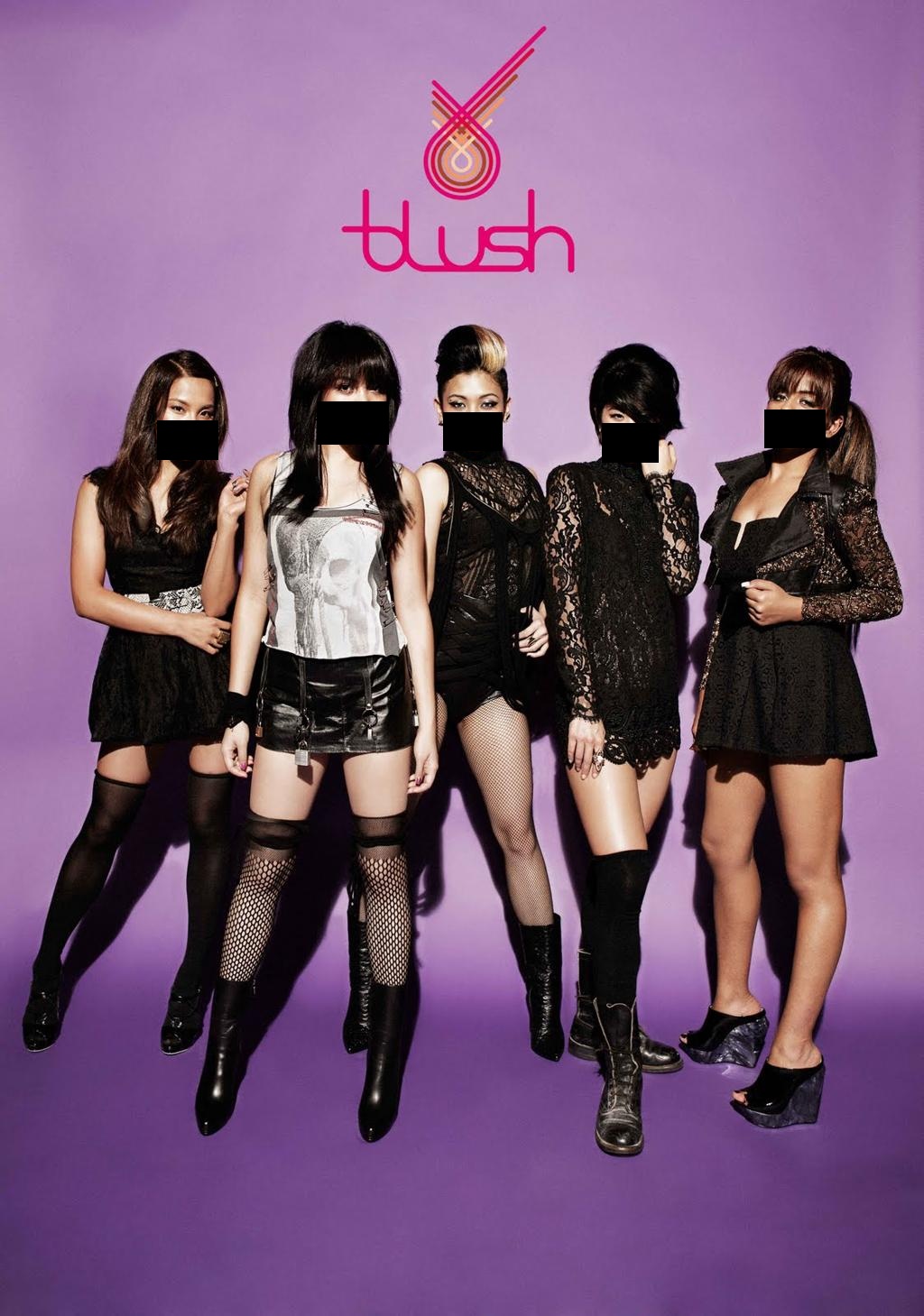}
    \caption{\label{fig:masked_sample_images}Some of the images obtained after blackening the lower half of the WIDER Face dataset \cite{widerface}.}
\end{figure}

The quality of this self-generated dataset involving the lower half of faces being blacked was acceptable for us to move forward and the final result of a huge majority of the images came out as expected. Some of these images are shown in Figure \ref{fig:masked_sample_images}.

We would have also liked to test the models on the WIDER Face testing dataset, but we did not have access to the ground truth labels of the bounding boxes for the testing dataset, which were required to modify the dataset. We ran the pre-trained models on the validation datasets, and the results were plotted using the evaluation code provided by the authors of WIDER Face.

\section{Implementation Details of the Models}\label{sec:implementationdetails}

\subsection{RetinaFace}

RetinaFace displays over $90\%$ accuracy on all three of the WIDER Face validation datasets. According to the authors of the model, the primary reasons for this increased accuracy are the following::

\begin{itemize}
	\item They manually annotate five facial landmarks on the dataset (both eyes, the tip of the nose, and the corners of the lips).
	\item They add a self-supervised mesh-decoder branch for predicting the 3D shapes of faces. This is done in parallel with the existing supervised branches.
\end{itemize}

\begin{figure}[!htbp]
	\centering
	\includegraphics[scale=0.25]{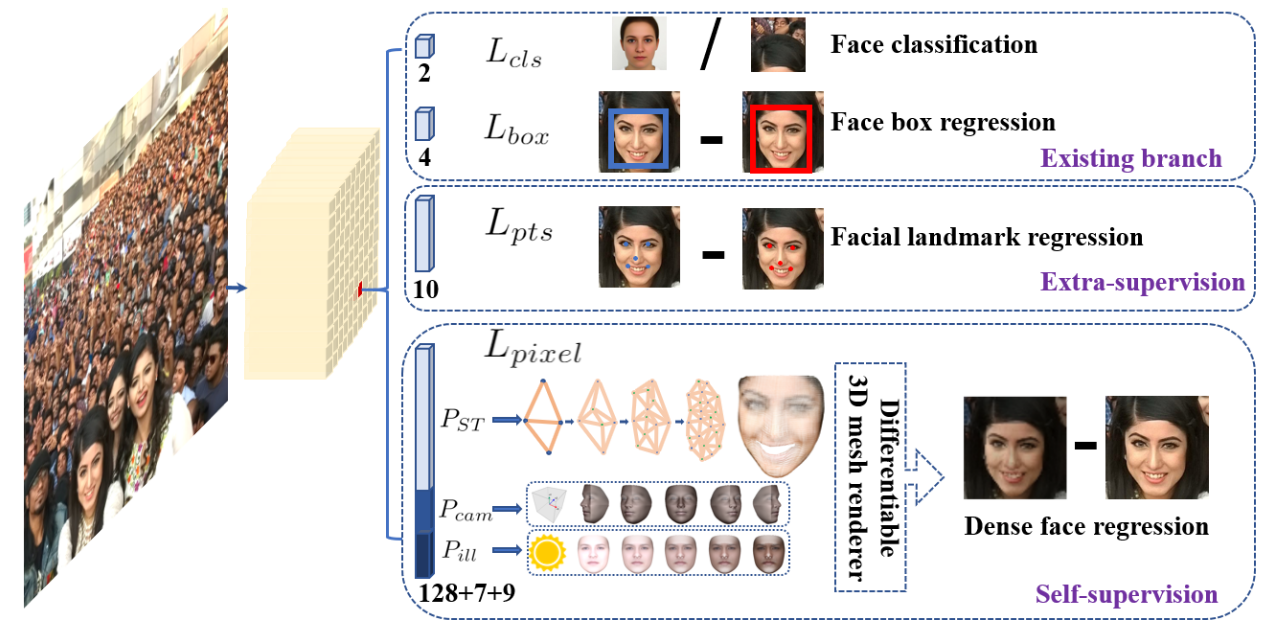}
	\caption{In RetinaFace, each positive anchor outputs (1) a face score, (2) a face box, (3) five facial landmarks, and (4) dense 3D face vertices projected on the image plane. Source: \cite{retinaface}.}
	\label{fig:retinaface}
\end{figure}

In addition, the following implementation details also play a crucial role in RetinaFace's high accuracy:

\begin{itemize}
	\item Large backbone: Starting from the most important contributor, RetinaFace uses Resnet-152 \cite{resnet} as its backbone, which one is of the largest neural network models when judged on the number of layers and trainable parameters.
	\item Utilizing contextual information: To better detect small and obscure faces, RetinaFace uses contextual information to locate them. This is done by predicting bounding boxes for human torsos simultaneously with those for faces and then using their relative positions to give confidence scores for faces.
	\item Filtering negative anchors: Another way in which RetinaFace captures small faces is by using very small anchors. This results in a large imbalance between positive and negative anchors, which will result in  reduced performance if not corrected. This model utilizes a method called online hard example mining (OHEM) \cite{ohem} to filter negative anchors and bring the ratio between negative and positive samples to at most 3:1.
	\item Data augmentation: Random square patches are cropped from the original images and resized to $640\times640$ to generate larger training faces. Other standard data augmentation techniques include random horizontal flip with $50\%$ probability and photo-metric color distortion.
\end{itemize}

Grounded on the extremely high accuracy displayed by RetinaFace, we expect it to perform well on the masked dataset as well. Of particular relevance here is RetinaFace's use of contextual information to locate faces, which would surely be an advantage in detecting masked faces. However, finding facial landmarks and face meshes on masked faces might be challenging, especially in cases where faces are small or already occluded, as in the case of the Hard validation set. We leave it to the results section to see how this might impact performance.

\subsection{Extremely Tiny Face Detector via Iterative Filter Reuse}

The EXTD has less than 0.1 million trainable parameters, which is indeed very tiny when compared with a model that uses a classification network like Resnet-152 with the number of parameters running into the tens and hundreds of millions \cite{extd}. This means that EXTD can be trained on devices with modest computing power while also making it deployable on edge devices such as smartphones and cameras for live detection. We chose EXTD to find out the accuracy achieved by lightweight models on the task of masked face detection.

EXTD's authors looked to create a model that is light enough to be able to run on CPU or mobile environments while still maintaining comparable accuracy to the deeper models. They achieve this by using what they call "Iterative Filter Reuse". In typical face detection algorithms that use convolutional layers, each convolutional layer is different with its own set of parameters. This means that increasing the number of layers would increase the complexity of the model, making it more taxing for devices to train and test. EXTD solves this problem by using a small backbone network to create the first feature map and then reusing the same network to create subsequent feature maps. This allows for an increase in the depth of the model without bloating the number of parameters.

Additional training strategies that EXTD employs are similar to RetinaFace, which include:
\begin{itemize}
	\item Balancing the ratio of positive and negative samples.
	\item Data augmentation such as random crop, color distortion, horizontal and vertical flips.
\end{itemize}

EXTD manages to get an accuracy of $82-86\%$ in the WIDER Face (Hard) dataset depending on the backbone network used \cite{extd}.

\subsection{A Light and Fast Face Detector for Edge Devices}

LFFD has a similar goal to EXTD, which is to create a face detector that can run on devices with limited storage and computing power such as mobile phones, CCTV cameras, and IoT (Internet of Things) sensors. This model seeks to balance both accuracy and running efficiency, which was achieved by reducing model complexity in mainly two ways - not using anchors, and using a CNN backbone with a small number of parameters.

\subsubsection{Receptive fields as natural anchors}

The best object detection models (of which face detectors are a subset) usually have pre-defined anchor boxes that are manually designed keeping in mind the range of sizes of objects that they expect to encounter. Hence these anchor boxes will be of varying sizes and aspect ratios in order to ensure maximal coverage. However, anchor-based methods come with three major challenges, as explained in \cite{lffd}:

\begin{itemize}
    \item It is difficult to cover all face scales with anchor matching.
    \item Anchors are matched to ground-truth bounding boxes by thresholding IOU (Intersection over Union) loss. This threshold is set empirically by trial and error, without any theoretical guidance.
    \item Choosing the number of anchors for different scales is also done via trial and error, and this may induce sample imbalance and redundant computation.
\end{itemize}

To circumvent these problems, the authors of LFFD argue that the receptive fields of neurons in feature maps can be considered to be inherent and natural 'anchors'. A receptive field is matched to a ground truth bounding box if and only if its center falls in the bounding box. Thus by eliminating the use of designated anchors, the authors of LFFD manage to eliminate considerable complexity from the model, resulting in a more simple and straightforward approach. With this implementation, LFFD can detect faces in the range of 10 to 560 pixels, matching those of several anchor-based strategies.

\subsubsection{Model Architecture}

The best performing models in the WIDER FACE dataset have either VGG16 \cite{vgg16} or Resnet50 as their backbones. LFFD on the other hand has its own novel CNN architecture consisting of only common layers (conv3x3, conv1x1, REctified Linear Unit (ReLU),  and residual connection). As a result, LFFD has only 2.1M parameters compared to 138.3M for VGG16 and 25.5M for ResNet50.

While LFFD and EXTD share the same goals, EXTD exhibits a much higher accuracy on the WIDER Face dataset. We imagine that this performance difference will also exist when testing on masked faces. No other specific implementation detail of these two models gives any clue as to how these models will perform on the masked dataset, although since both of these have much smaller backbone networks, we do not expect them to be able to match the accuracy achieved by RetinaFace.

\subsection{Multi-task Cascaded Convolutional Networks}
The authors of MTCCN sought to introduce the importance of the correlation between face detection and face alignment methods and improve upon the existing models that did the same. They also tackled the issues with manual hard sample mining by automating this process through an online hard sample mining method. To this effect, the authors proposed a new lightweight cascaded CNN-based framework for joint face detection and alignment to tackle the two objectives.

The algorithm works in the following four steps:
\begin{enumerate}
    \item Form an image pyramid by resizing the input images to different scales.
    \item Use a CNN called the Proposal Network (P-Net) to obtain candidate facial boxes and bounding box regression vectors, followed by calibration and merging of highly overlapped candidates.
    \item Use another CNN called the Refine Network (R-Net) to reject a large number of false candidates, and then again followed by calibration and a merging of highly overlapped candidates.
    \item Finally, use a CNN called the Output Network (O-Net) to identify the positions of 5 facial landmarks.
\end{enumerate}

\begin{figure*}
    \centering
    \includegraphics[scale=0.3]{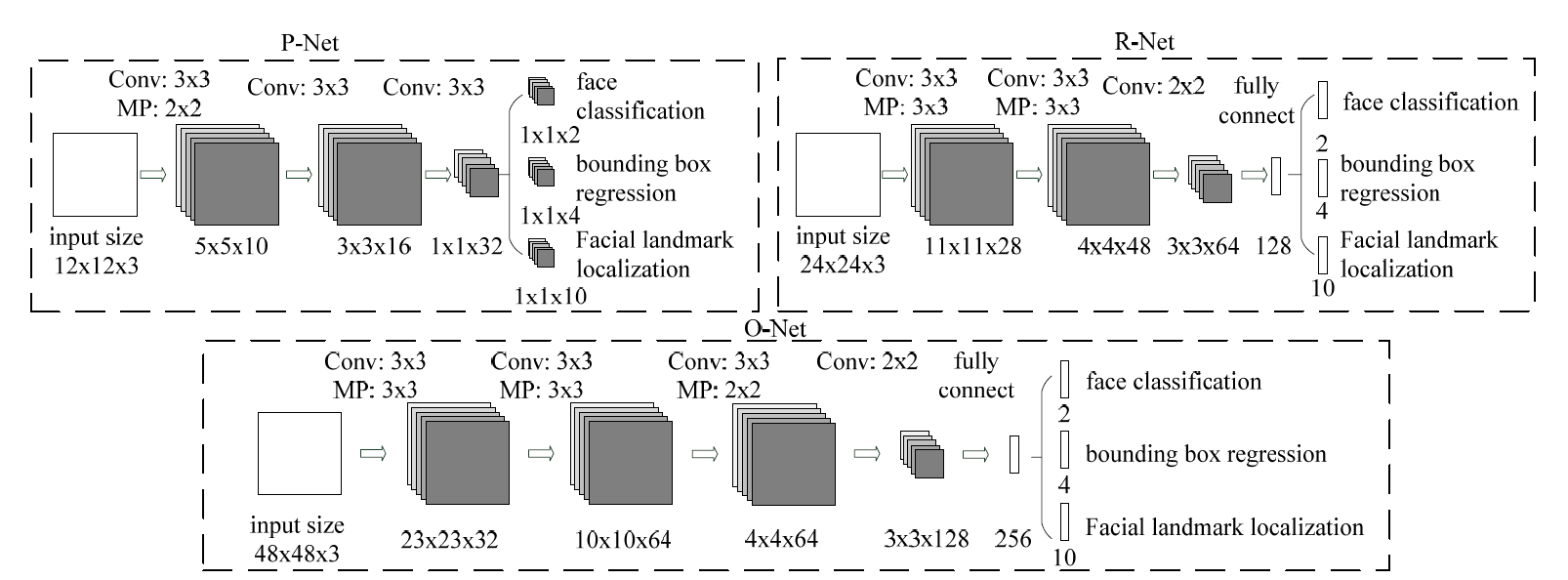}
    \caption{The architectures of P-Net, R-Net, and O-Net are used in \cite{mtccn}. Here “MP” and “Conv” stands for max pooling and convolution respectively. The step size in convolution and pooling are 1 and 2, respectively. Source: \cite{mtccn}}
    \label{fig:mtccn_model}
\end{figure*}

The CNN architectures used in this model are as shown in Figure \ref{fig:mtccn_model}. The authors also proposed an online hard sample mining to help strengthen the model during training. This method is adapted to the training process and involves sorting the losses computed for each mini-batch during forward propagation and tagging the top 70\% of them as hard samples. They compute the gradients only from these hard samples in the backward propagation. The authors have shown that using this process helps them obtain better performance.

Another important aspect the authors wanted to cover was to reduce the computation while increasing the depth to get better performance. They improved upon CNN architectures \cite{ref19} by reducing the number of filters and changing 5$\times$5 filters to 3$\times$3 filters. They felt that face detection might need a lesser number of filters per layer as unlike other multiclass-classification tasks, face detection is a binary problem \cite{mtccn}. With these improvements, compared to the previous architecture in \cite{ref19}, the authors obtained better performance with lesser runtime.

\subsection{CenterFace}

\begin{figure}[htpb]
    \centering
    \includegraphics[scale=0.6]{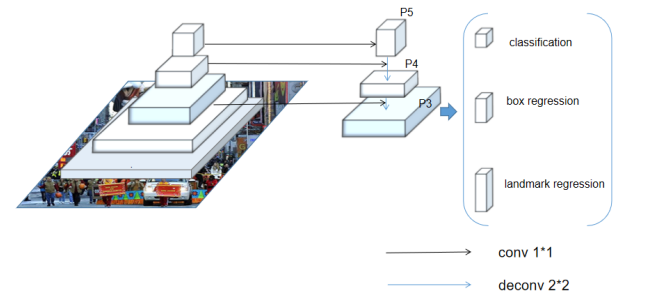}
    \caption{The architecture of the CenterFace model. Source: \cite{centerface}}
    \label{fig:centerface_arch}
\end{figure}

The authors of CenterFace designed the model to make it light and yet powerful by proposing a detection and alignment method that has greater efficacy than other anchor-free object detection frameworks \cite{ref1, ref3, ref6, ref14, ref15, FPN, ref26}. The architecture of CenterFace as shown in Figure \ref{fig:centerface_arch}, and the model can be trained end-to-end. It also does not have an incredibly large backbone like VGG16 that most state-of-the-art models usually use, and instead Mobilenetv2 \cite{mobilenetv2} is used as the backbone. 

The model runs according to the following workflow:
\begin{enumerate}
    \item The input is taken in a single scale and passed through a feature pyramid built using the Feature Pyramid Network \cite{FPN}.
    \item The face classification branch generates a heatmap Y, where a face center will correspond to Y$_{x,y}$ = 1 and the background will correspond to Y$_{x,y}$ = 0. 
    \item Size and landmarks of a face are determined from image features of the center locations through regression.
    \item Although face classification is done at different scales of the input image, facial size and landmark determination is done only at one scale, to reduce computational requirements. 
\end{enumerate}

\subsection{FaceBoxes}
\begin{figure*}[htpb]
    \centering
    \includegraphics[scale=0.4]{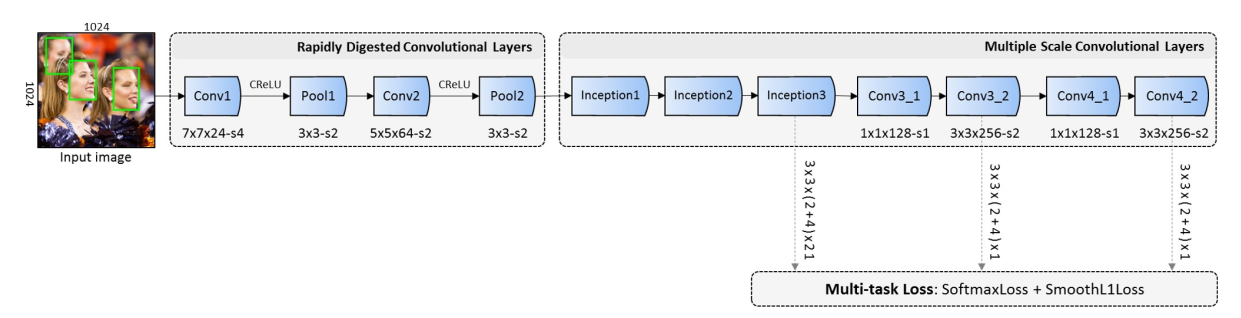}
    \caption{Faceboxes' Network Architecture. Source: \cite{faceboxes}}
    \label{fig:facebox-arch}
\end{figure*}

Using face detectors with a CPU that can demonstrate a good real-time speed and high performance has been a major concern in this field for a long time and FaceBoxes looks to address this 2 pronged problem. The speed being independent of the number of faces in an image is also a key feature of this model. The model is divided into three sections, the Rapidly Digested Convolutional Layer (RDCL), the Multiple Scale Convolutional Layer (MSCL), and the anchor densification strategy. The network architecture is as shown in Figure \ref{fig:facebox-arch}.

Convolutions in any network for large input images are computationally heavy given the number of arithmetic operations involved. RDCL looks to reduce the size of the inputs rapidly (by setting large stride lengths for convolutional and pooling layers) to help achieve faster run-time on CPUs. As we can see in Figure \ref{fig:facebox-arch}, the stride lengths of Conv1, Pool1, Conv2, and Pool2 layers are 4, 2, 2, and 2 respectively, which is equivalent to reducing the spatial size of the input image by a factor of 32. Another thing to note is that RDCL uses Concatenated ReLU \cite{crelu} which also plays an important role in increasing speed with a negligible decrease in accuracy.

\begin{figure}[h]
    \centering
    \includegraphics[width = 2in]{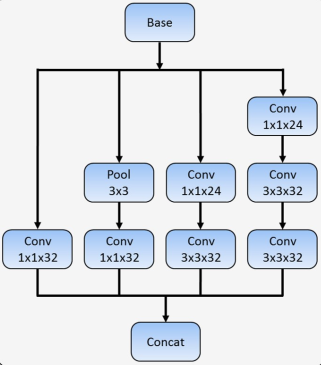}
    \caption{The Inception module. Source: \cite{faceboxes}.}
    \label{fig:inception}
\end{figure}

MSCL is a method that is conceptually similar to Region Proposal Networks (RPNs) \cite{rpn}. RPN involves associating anchors with only the last convolutional layer, thereby making it difficult to handle faces of different scales. To address this, MSCL goes with a new anchor strategy and an architecture making it suitable to detect faces of various scales. The progressive decrease in the size of layers throughout MSCL helps employ this. These layers also discretize anchors over multiple layers with different resolutions to handle faces of various sizes.  From the MSCL architecture shown in Figure \ref{fig:facebox-arch}, we see that there are inception layers at the beginning which are effective algorithms to capture faces of different scales. Inception modules themselves are computationally costly but inculcating 1$\times$1 convolutional layers help drastically bring down the number of arithmetic operations required to get a similar output. The inception module implementation is as shown in Figure \ref{fig:inception}.

The anchor densification strategy increases the density of small anchors to improve the recall rate of small faces. For the anchor size, the authors have imposed a 1:1 aspect ratio (i.e., square anchor), because the face box is approximately taken to be a square. The size of anchors and tiling intervals are different for different layers. The tiling interval of an anchor on the image is equal to the stride size of the corresponding anchor-associated layer. The anchor densification strategy thus aims to densify one type of anchor n times, i.e, it uniformly tiles n$^2$ anchors around the center of one receptive field instead of the standard procedure of one box at the center of this receptive field.

Some training strategies employed by the authors were:
\begin{itemize}
    \item Each training image went through data augmentation such as color distortion, random cropping, scale transformation, horizontal flipping, and face-box filters.
    \item A matching strategy involving Jaccard overlap of the best matching anchor with a face, and also anchors to any face are higher than a threshold.
    \item Hard negative mining.
\end{itemize}

As stated in \cite{faceboxes}, the model can achieve up to 20FPS for VGA-resolution images on a CPU with state-of-the-art accuracy, and the authors tested for speed using a Titan X (Pascal) GPU, and cuDNN v5.1 with an Intel Xeon E5-2660v3@2.6GHz processor. To understand how each component of the model contributes to the accuracy, ablative-based tests were done using the FDDB database \cite{fddb}, the results of which are shown in Table \ref{tab:faceboxes-ablation-results}. 

\begin{table}[htpb]
\centering
\begin{tabular}{|c|cccc|}
\hline
Contributions                 & \multicolumn{4}{c|}{FaceBoxes}         \\ \hline
RDCL                          &       &          &          & $\times$ \\ \hline
MSCL                          &       &          & $\times$ & $\times$ \\ \hline
Anchor densification strategy &       & $\times$ & $\times$ & $\times$ \\ \hline
Accuracy (mAP)                & 96.0  & 94.9     & 93.9     & 94.0     \\ \hline
Speed (ms)                    & 50.98 & 48.27    & 48.23    & 67.48    \\ \hline
\end{tabular}
\caption{\label{tab:faceboxes-ablation-results}Ablative test results. Source: \cite{faceboxes}}.
\end{table}

From Table \ref{tab:faceboxes-ablation-results}, we see that removing the anchor densification strategy led to a decline in mAP by 1.1\%. Removing MSCL and the anchor densification strategy led to a decline of 1.0\% in accuracy in comparison to only the anchor densification strategy being removed. RDCL also was seen to contribute to a significant increase in speed with an insignificant decrease in accuracy (mAP decreases by 0.1\%).

\subsection{TinyFaces}

TinyFaces concerns itself with the explicit task of detecting small faces, which it addresses primarily through the methods discussed in the following subsections:.

\subsubsection{Scale-specific detectors}
The majority of face detection algorithms, including the other ones considered in this study, are scale-invariant. ie, they have a single detection pipeline for detecting faces of all scales. TinyFaces deviates from this norm by using separate detectors for different scales and aspect ratios. The authors argue that this is needed because in a finite resolution image, the cues for recognizing a 300 by 300 pixel image are very different from those required to detect a 3 by 3 pixel image \cite{tinyfaces}. In order to alleviate the problems that come with using different detectors simultaneously such as a lack of training data for individual scales and the computational cost of running a large number of detectors at the same time, the authors train and run scale-specific detectors in a multi-task fashion: the detectors receive their inputs from different layers of the same feature hierarchy (backbone network).

\subsubsection{Context}

Similar to some of the other face detection models that we looked at previously, TinyFaces also considers contextual information for detecting faces. The authors conducted several experiments to show that the use of context boosts accuracy in detecting faces of any scale, but the increments obtained in detecting small faces are much higher than those obtained for larger faces. To find contextual information, they use \textit{massively-large receptive fields}, where the faces to be detected occupy less than $1\%$ of the templates' sizes \cite{tinyfaces}.

\begin{figure*}[htbp]
	\centering
	\includegraphics[scale=0.6]{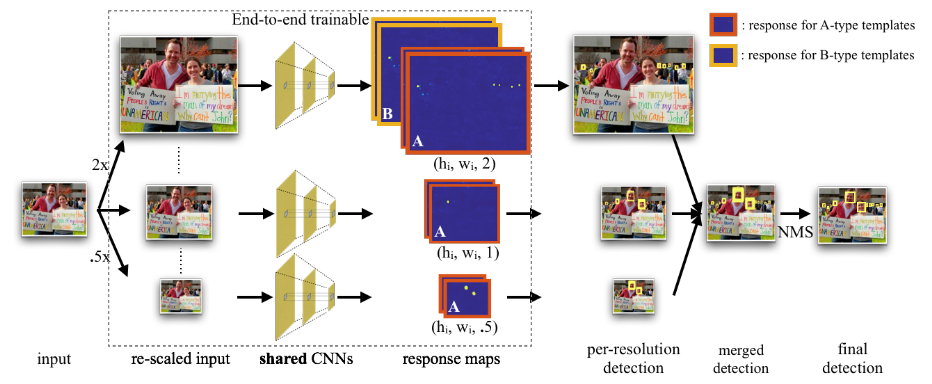}
	\caption{The architecture of TinyFaces. The outputs from the models at different resolutions are combined using non-maximum suppression (NMS) at the original resolution to get the final results \cite{tinyfaces}.}
	\label{fig:tinyfaces}
\end{figure*}

\subsubsection{Resolution}

TinyFaces uses a ResNet-50 backbone that has been pre-trained on the ImageNet \cite{imagenet} object-detection dataset. As part of several experiments, the authors found that using a 50$\times$40 pixel template for detecting 25$\times$20 pixel faces by up-sampling the images to twice their size had the effect of boosting performance. A similar increase in performance was also seen when using the same template for detecting down-sampled, larger faces. The authors traced the root of this observation to the distribution of scales in the pre-training dataset, which was ImageNet in their case. More than $80\%$ of object categories in the ImageNet dataset have an average size in the range of 40-140 pixels \cite{tinyfaces}, and this imbalance in training data explains why receptive fields of certain sizes yield greater performance than others. This is an important observation with great implications for existing face detection models as almost all of them use a backbone network that has been pre-trained on some object-detection dataset.

The resolution up-sampling method mentioned above is very effective in boosting the accuracy of detecting small faces. This will be particularly beneficial for the Hard validation dataset because it has a large percentage of tiny faces. This advantage may be amplified in the case of the masked dataset because now in effect faces are half their previous sizes (since the lower half of each face is blacked).

\section{Results and Evaluation}\label{sec:results}

\begin{figure}[htpb]
     \centering
     \begin{subfigure}[b]{0.49\textwidth}
         \centering
         \includegraphics[width=\textwidth]{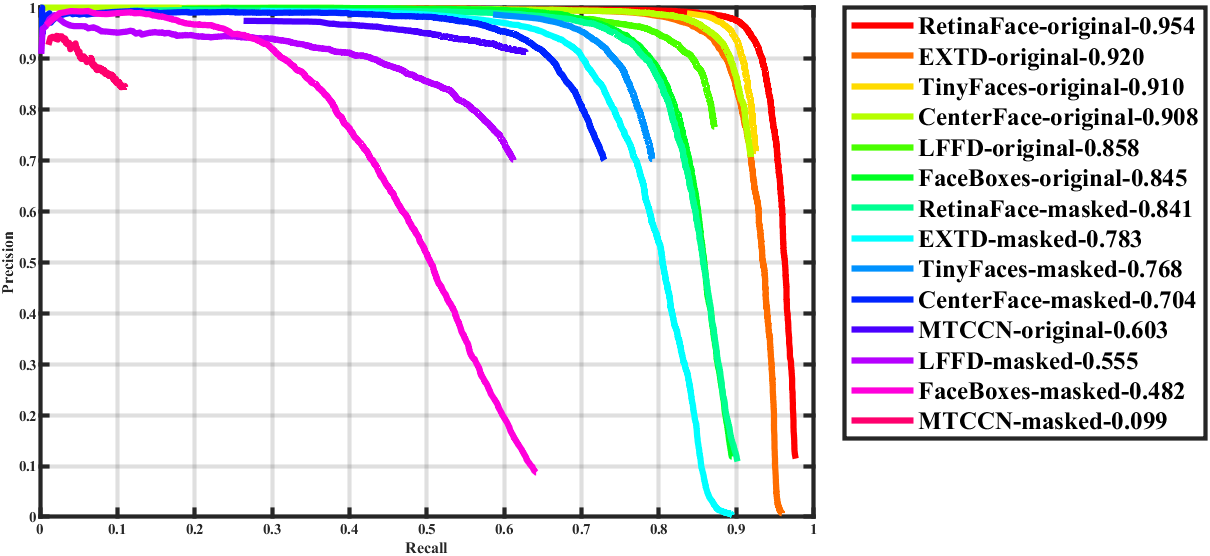}
         \caption{}
         \label{fig:easy_val_results}
     \end{subfigure}
     \hfill
     \begin{subfigure}[b]{0.49\textwidth}
         \centering
         \includegraphics[width=\textwidth]{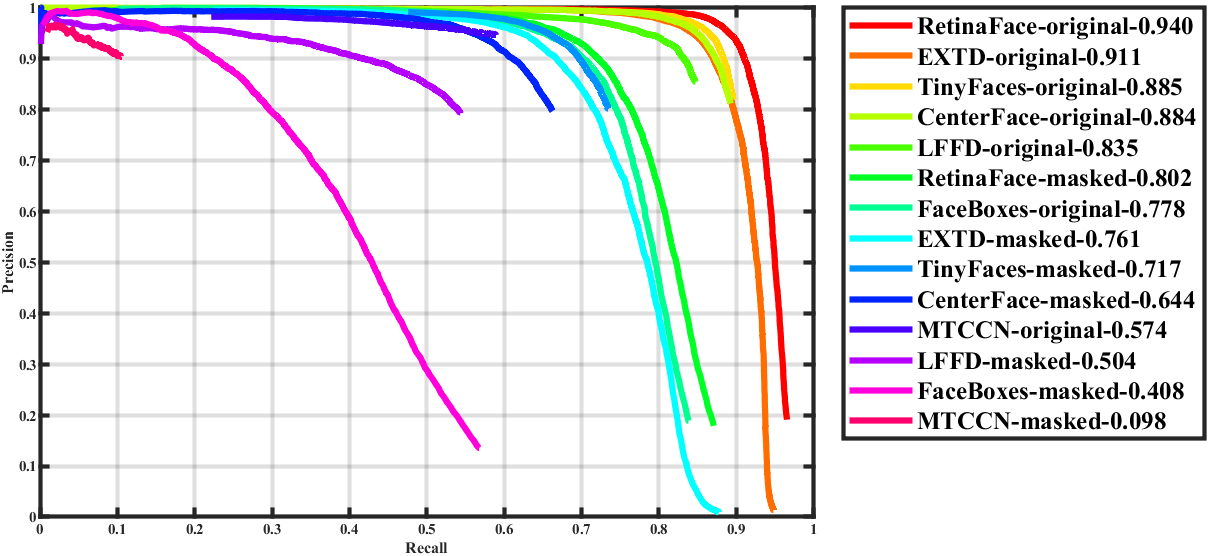}
         \caption{}
         \label{fig:med_val_results}
     \end{subfigure}\newline
     \hfill
     \begin{subfigure}[b]{0.5\textwidth}
         \centering
         \includegraphics[width=\textwidth]{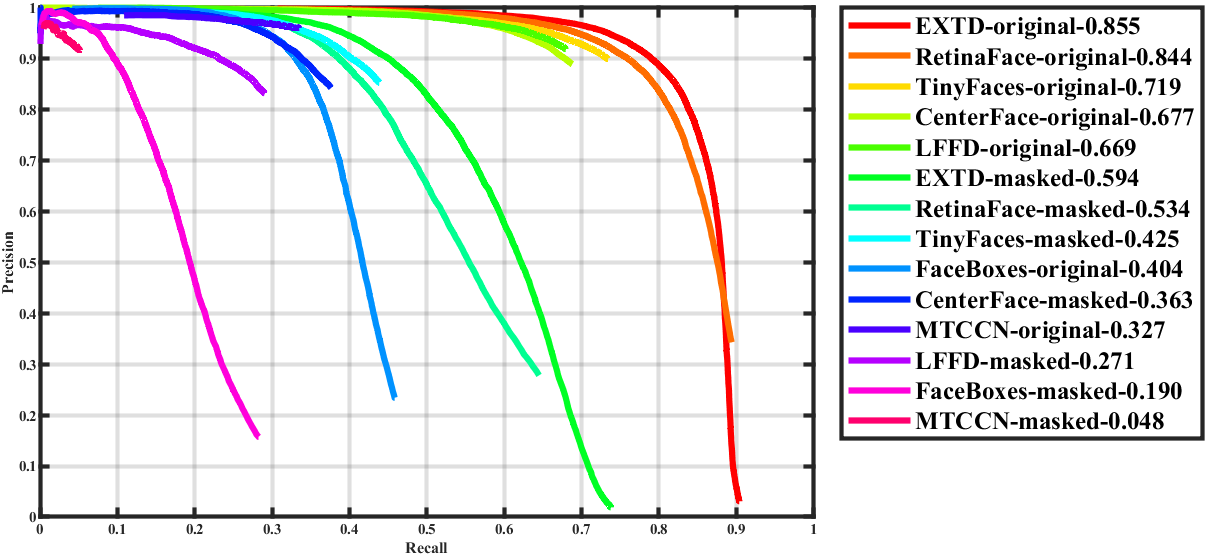}
         \caption{}
         \label{fig:hard_val_results}
     \end{subfigure}
    \caption{PR plots obtained for (a) Easy, (b) Medium, and (c) Hard validation datasets. The suffixes original and masked imply that the model accuracy was calculated for the original and masked datasets respectively.}
    \label{fig:val_results}
\end{figure}

The plots obtained for the results of the various models that we ran are given in Figure \ref{fig:val_results}. These plots include the accuracy of the models when run on the unaltered WIDER Face dataset as well as the blacked datasets. Note that the accuracies reported for the models when run on the unaltered dataset are what we obtained when running the models on our machine, and not what the authors of the various models reported. Overall, we see that in detecting masked faces, RetinaFace does best on the Easy and Medium validation datasets, but on the Hard dataset, EXTD comes on top. The best accuracies reported are $84\%$ on the Easy validation dataset, $80\%$ on Medium (both by RetinaFace), and $59\%$ on Hard (by EXTD).

Let us look at how the models rank with respect to each other for the different validation datasets. Looking at Table \ref{tab:ranking-of-models}, we see that most models (serial numbers 3-5 and 10-14) have retained their overall position across all three subsets. As we compare the Easy and Medium sets, we see that M-RetinaFace improves performance, doing better than O-Faceboxes. And from Medium and Hard sets, O-EXTD performs better than O-RetinaFace and M-EXTD goes up by 2 positions and O-FaceBoxes does worse than M-TinyFaces. Other models like MTCCN and LFFD remain at the same ranks.

The models that witnessed a change in their relative performances as one goes from the Easy to Hard datasets are RetinaFace, EXTD, FaceBoxes, and TinyFaces. We will now look at why these changes may have taken place, and what they can teach us about the requirements to accurately detect masked faces.

\begin{table*}[htpb]
\centering
\begin{tabular}{|p{1cm}|p{3cm}|p{3cm}|p{3cm}|}
\hline
Sl. No. & Easy dataset & Medium datasat & Hard dataset \\ \hline
1       & O$^*$-Retinaface & O-Retinaface   & O-EXTD       \\ \hline
2       & O-EXTD       & O-EXTD         & O-Retinaface \\ \hline
3       & O-TinyFaces  & O-TinyFaces    & O-TinyFaces  \\ \hline
4       & O-CenterFace & O-CenterFace   & O-CenterFace \\ \hline
5       & O-LFFD       & O-LFFD         & O-LFFD       \\ \hline
6       & O-FaceBoxes  & M$^\#$-RetinaFace   & M-EXTD       \\ \hline
7       & M-RetinaFace & O-FaceBoxes    & M-RetinaFace \\ \hline
8       & M-EXTD       & M-EXTD         & M-TinyFaces  \\ \hline
9       & M-TinyFaces  & M-TinyFaces    & O-FaceBoxes  \\ \hline
10      & M-CenterFace & M-CenterFace   & M-CenterFace \\ \hline
11      & O-MTCCN      & O-MTCCN        & O-MTCCN      \\ \hline
12      & M-LFFD       & M-LFFD         & M-LFFD       \\ \hline
13      & M-FaceBoxes  & M-FaceBoxes    & M-FaceBoxes  \\ \hline
14      & M-MTCCN      & M-MTCCN        & M-MTCCN      \\ \hline
\multicolumn{4}{@{}l}{$^*$M--Model tested in the masked dataset\qquad 
$^{\#}$ O--Model tested in the original dataset.}
\end{tabular}%
\caption{Models ranked based on performance for different WIDER Face validation datasets.}
\label{tab:ranking-of-models}
\end{table*}

For RetinaFace, based on ablation studies done by the model's authors, it is safe to conclude that the superior accuracy of RetinaFace (for both regular and masked faces) is primarily due to its extensive backbone network. The two critical additions for RetinaFace compared to the other models are facial landmark localization and dense regression. Even without including these two features, the authors report an accuracy of $91.286\%$ on the Hard validation dataset, making it by far the best performing model among those that we have considered, a fact that we have verified. Surprisingly, on the Hard masked dataset, we see that EXTD eclipses RetinaFace. In the ablation studies done by RetinFace's authors, they report that when adding the dense regression branch, accuracy increases on the Easy and Medium subsets while it decreases on the Hard subset. They attribute this to the Hard dataset having a much more significant proportion of small and occluded faces, making dense regression challenging. Although they noted that using landmark localization and dense regression together boosted accuracy on all three validation subsets, it appears probable that for the Hard masked dataset, learning facial landmarks cannot offset the dip caused by dense regression, leading to a loss in performance.

TinyFaces goes up a rung on the ranking for the Hard validation dataset, which can also be explained based on the implementation details that we looked at in Section \ref{sec:implementationdetails}. TinyFaces scales up small faces and scales down large faces to bring them to a preferred range, which is identified based on the distribution of image sizes in the pre-training dataset. The benefits of this method will be most evident when the dataset contains a significant number of challenging images in which faces may be small, poorly illuminated, or otherwise occluded, a description that aptly summarizes the Hard dataset.

Faceboxes apart from MTCCN, is the worst-performing. It proposed a fast and high-performing detector and had three important features: RDCL, MSCL, and the anchor densification strategy. RDCL does not focus on improving accuracy, it only helps in achieving faster run-times. This can be seen from the ablative tests. To tackle performance improvement, the authors looked at only being able to detect faces at different scales. But there are other factors that inhibit performance like occlusion, faces at different angles (poses), illumination, etc. MSCL goes with a new anchor strategy and architecture. The progressive decrease in the size of layers throughout MSCL helps employ the detection of different scaled faces. These layers also discretize anchors over multiple layers with different resolutions of handle faces of various sizes). Although the ablative tests show that the MSCL and anchor densification strategy only contribute to 2.1\% of the total accuracy, FaceBoxes does not have an architecture tailored to improve accuracies, let alone perform well for occluded faces.





\section{Conclusions}\label{sec:conclusions}

We found that of the various face detection algorithms that we tested, which was a diverse and fully representative sample of current state-of-the-art models, the best ones yielded an accuracy of around $80\%$ on the Easy and Medium (masked) validation datasets. This shows that the state-of-the-art face detection models already show decent accuracies in detecting masked faces, which is all the more impressive when considering the fact that these models were trained using a normal (original) dataset. On the Hard (masked) validation dataset, however, the best model showed only about $60\%$ accuracy, which is inadequate for most practical applications.

The reason that these models performed badly on the Hard validation dataset can be attributed to this dataset having a relatively large proportion of small and partially occluded faces. On top of that, our testing dataset had $50\%$ of the area of each face blacked out, which will have left very little information for the models to process, thereby resulting in a far lower inference accuracy. However, we believe that this dropoff in accuracy can be offset if the models were trained on a masked dataset, which should boost the accuracy across all three validation datasets. In fact, that is the greatest current limitation, the lack of a proper masked face dataset that is large enough to encompass the true diversity that we can find in the wild. The way that this study was structured, the models were trained on normal faces and were then used to detect faces with half the area missing. Hence, in effect, the models had only half the information to go on with during the testing phase. This is why we believe that the results obtained in this review are only a lower bound on the masked face detection accuracies achievable by these models, as using more appropriate datasets for training and testing, as and when they become available, are sure to increase their performance. Seen in this light, the results of this study are encouraging, as $80\%$ accuracy on the Easy and Medium (masked) validation datasets without any masked face specific modifications whatsoever attests to the incredible progress that face detection models, and more generally, object detection models have made in the past few years.

\begin{figure}[h]
     \centering
     \begin{subfigure}[b]{0.3\textwidth}
         \centering
         \includegraphics[width=\textwidth]{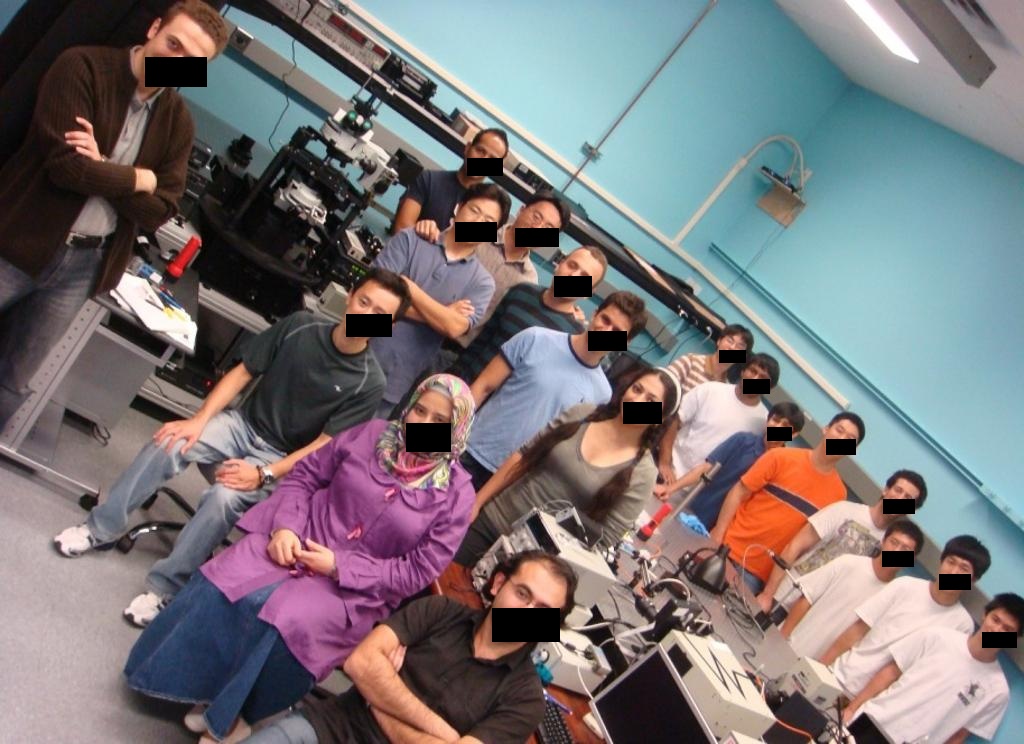}
         \caption{}
         \label{fig:bad_masked_img1}
     \end{subfigure}
     \hfill
     \begin{subfigure}[b]{0.3\textwidth}
         \centering
         \includegraphics[width=\textwidth]{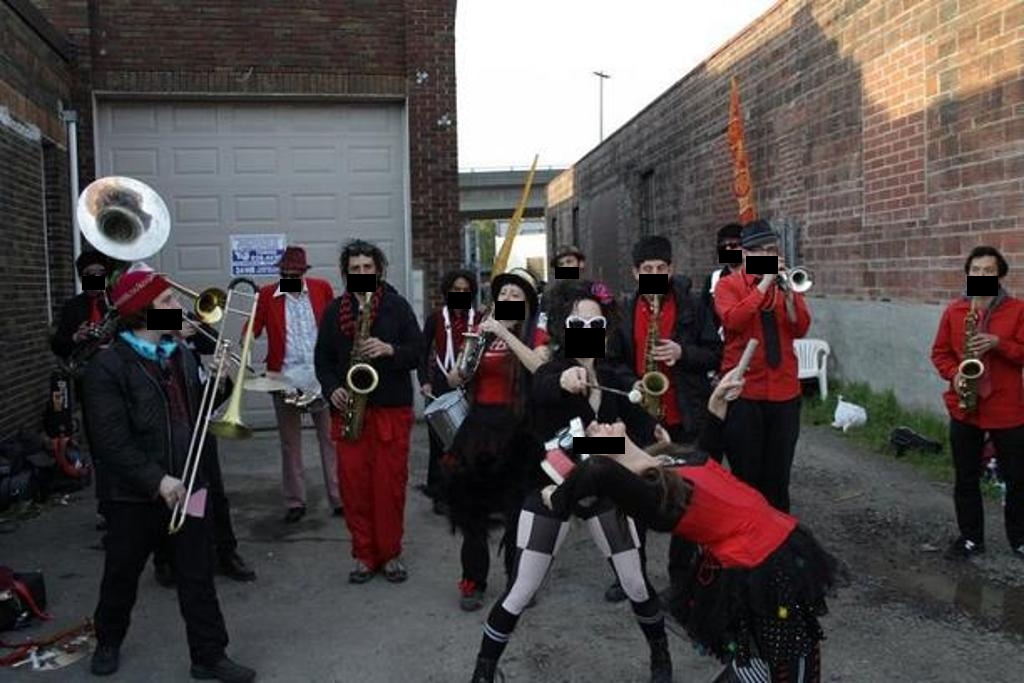}
         \caption{}
         \label{fig:bad_masked_img2}
     \end{subfigure}
     \hfill
     \begin{subfigure}[b]{0.3\textwidth}
         \centering
         \includegraphics[width=\textwidth]{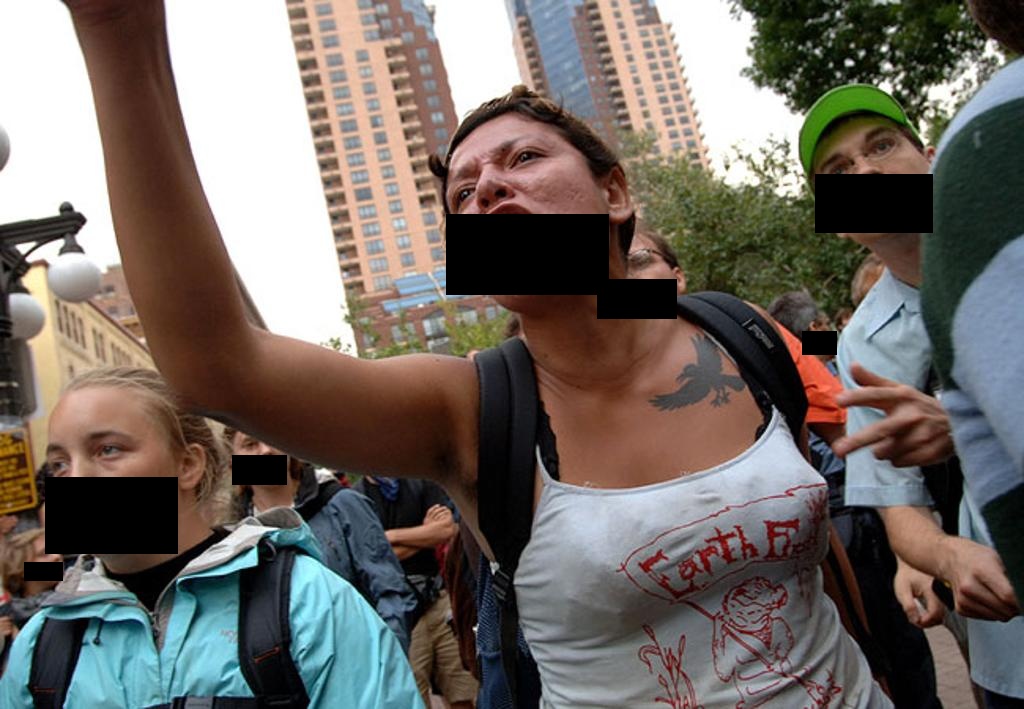}
         \caption{}
         \label{fig:bad_masked_img3}
     \end{subfigure}
     \hfill
    \caption{Instances of poorly simulated images obtained after blackening the lower half of the WIDER Face dataset.}
    \label{fig:bad_masked_sample_images}
\end{figure}

Although we were able to obtain some goof images from the self-generated masked dataset, there were some (minor number of) outliers that had to be noted which is not a major concern, but yet a limitation to our testing process. This is self-explanatory from Figure \ref{fig:bad_masked_sample_images}, where we see two kinds of problems from these images, namely that the orientation of faces matter, and the closeness of 2 faces also matter. Figures \ref{fig:bad_masked_img1} and \ref{fig:bad_masked_img2} (see bottom-most face) is an instance of the scenario explained in the first point and Figure \ref{fig:bad_masked_img3} conveys the second point.

\bibliographystyle{ieeetr}
\bibliography{references}

\end{document}